\documentclass[compsoc,11 pt,onecolumn, conference]{IEEEtran}
\IEEEoverridecommandlockouts                              % This command is only
\usepackage{algorithm2e}
\ifCLASSOPTIONcompsoc
  \usepackage[nocompress]{cite}
\else
  \usepackage{cite}
\fi
\ifCLASSINFOpdf
\else
\fi

\usepackage[]{footmisc}

\usepackage{comment}
\usepackage{multirow}
\usepackage{amsfonts}
\usepackage{booktabs}
\usepackage{siunitx}
\usepackage{placeins}
\usepackage{epsfig} % for postscript graphics files
\usepackage{mathptmx} % assumes new font selection scheme installed
\usepackage{amsmath} % assumes amsmath package installed
\usepackage{amssymb}  % assumes amsmath package installed
\usepackage{balance}
\usepackage{authblk}
\usepackage{url}

\title{%% \LARGE \bf
Transfer Learning for Microstructure Segmentation with CS-UNet: A Hybrid Algorithm with Transformer and CNN Encoders
}

\begin{document}

\author{
Khaled Alrfou\\
U. of Wisconsin -- Milwaukee \\
kalrfou@uwm.edu
\and
Tian Zhao \\
U. of Wisconsin -- Milwaukee \\
tzhao@uwm.edu
\and 
Amir Kordijazi \\
U. of Southern Maine \\
Amir.kordijazi@maine.edu
%% kordija2@uwm.edu 
} 

\maketitle
\thispagestyle{plain}
\pagestyle{plain}

%%%%%%%%%%%%%%%%%%%%%%%%%%%%%%%%%%%%%%%%%%%%%%%%%%%%%%%%%%%%%%%%%%%%%%%%%%%%%%%%
\begin{abstract}
Transfer learning improves the performance of deep learning models by initializing them with parameters pre-trained on larger datasets. Intuitively, transfer learning is more effective when pre-training is on the in-domain datasets. A recent study by NASA has demonstrated that the microstructure segmentation with encoder-decoder algorithms benefits more from CNN encoders pre-trained on microscopy images than from those pre-trained on natural images. However, CNN models only capture the local spatial relations in images. In recent years, attention networks such as Transformers are increasingly used in image analysis to capture the long-range relations between pixels. In this study, we compare the segmentation performance of Transformer and CNN models pre-trained on microscopy images with those pre-trained on natural images. Our result partially confirms the NASA study that the segmentation performance of out-of-distribution images (taken under different imaging and sample conditions) is significantly improved when pre-training on microscopy images. However, the performance gain for one-shot and few-shot learning is more modest with Transformers. We also find that for image segmentation, the combination of pre-trained Transformers and CNN encoders are consistently better than pre-trained CNN encoders alone. Our dataset (of about 50,000 images) combines the public portion of the NASA dataset with additional images we collected. Even with much less training data, our pre-trained models have significantly better performance for image segmentation. This result suggests that Transformers and CNN complement each other and when pre-trained on microscopy images, they are more beneficial to the downstream tasks.
\end{abstract}

%%%%%%%%%%%%%%%%%%%%%%%%%%%%%%%%%%%%%%%%%%%%%%%%%%%%%%%%%%%%%%%%%%%%%%%%%%%%%%%%
\section{Introduction}

%% outline of this section
%% 1. background of transfer learning and why it is useful to pretain on in-domain data
%% 2. what does prior work did -- NASA paper
%%    (a) pretained on micronet images
%%    (b) CNN trained on classification is used to initialize encoder of the segmentation network
%%    (c) result is better for one-shot, few-shot, and out of distribution segementation
%%    core-premise: first few convolution kernels are relevant since they capture lower level features while deeper kernels capture high-level information that is not relevant to microscopy images
%% 3. what do we do
%%    (a) CNN only captures local relations between pixels. Attention network based transformers capture long-distance relations between pixels. Conjecture is that the combined networks in transfer learning should provide superior results
%%    (b) CNN is used to initialize the encoder but Swin-T is used to initialize both the encoder and the decoder
%%    (c) results show that the improvement on one-shot/few-shot learning using in-domain images is more modest while the improvement on out-of-distribution images is similar
%%    (d) results are superior than micronet despite having half the number of pre-training images

Microscopy imaging provides real information about matters, but retrieving quantitative information on morphology, size, and distribution requires manual measurements on micrographs, which is not only time-consuming and labor-intensive but also prone to biases. The length and time scales of material structures and phenomena differ significantly among the components, which adds to the complexity. Creating connections between process, structure, and properties is thus a challenging problem~\cite{stuckner2022microstructure,ge2020deep}.

Deep Learning (DL) has been widely applied to complex systems because of its ability to extract important information automatically. Researchers have applied DL algorithms to image analysis to identify structures and to determine the relationship between microstructure and performance. 
DL has been demonstrated to complement physics-based methods for materials design. However, DL requires large amount of training data while the limited number of microscopy images tends to reduce its effectiveness. Learning techniques, such as transfer learning, multi-fidelity modeling, and active learning, were developed to make DL applicable to smaller datasets~\cite{ge2020deep,choudhary2022recent}.
Transferring learning uses the parameters of a model pre-trained on a larger dataset to initialize a model trained on a smaller dataset for a downstream task. For example, a Convolutional Neural Network (CNN) pre-trained on natural images can be used to initialize a neural network for image segmentation to improve its precision and reduce the training time.   

Pre-training with natural images such as ImageNet is not ideal since the models trained on natural images identify high-level features that do not exist in microscopy images. Recent work by Stuckner {\em et al.}~\cite{stuckner2022microstructure} demonstrated the advantage of pre-training CNN with a microscopy dataset called MicroNet that has over 110,000 images. 
They evaluated pre-trained CNN encoders for segmenting microscopy images of nickel-based superalloys (Super) and environmental barrier coatings (EBCs). Pre-training with MicroNet resulted in significantly higher accuracy measured in {\em Intersection over Union} (IoU) for one-shot and few-shot learning and for out-of-distribution images that have different compositions, etching, and imaging conditions than the training images.

In recent years, attention-based neural networks called Transformers are widely adopted in computer vision. 
While CNN extracts features from local regions of images using convolution filters to capture the spatial relation between the pixels, Transformer divides an image into patches and feeds them into a Transformer-based encoder to capture the long-range relation between pixels across the images~\cite{dosovitskiy2020image,alrfou2022computer}.
Thus, it is possible that a combination of CNN and Transformer may be more effective in transfer learning than either of the models alone.

\begin{figure}[ht]
\centering
  \includegraphics[width=175mm]{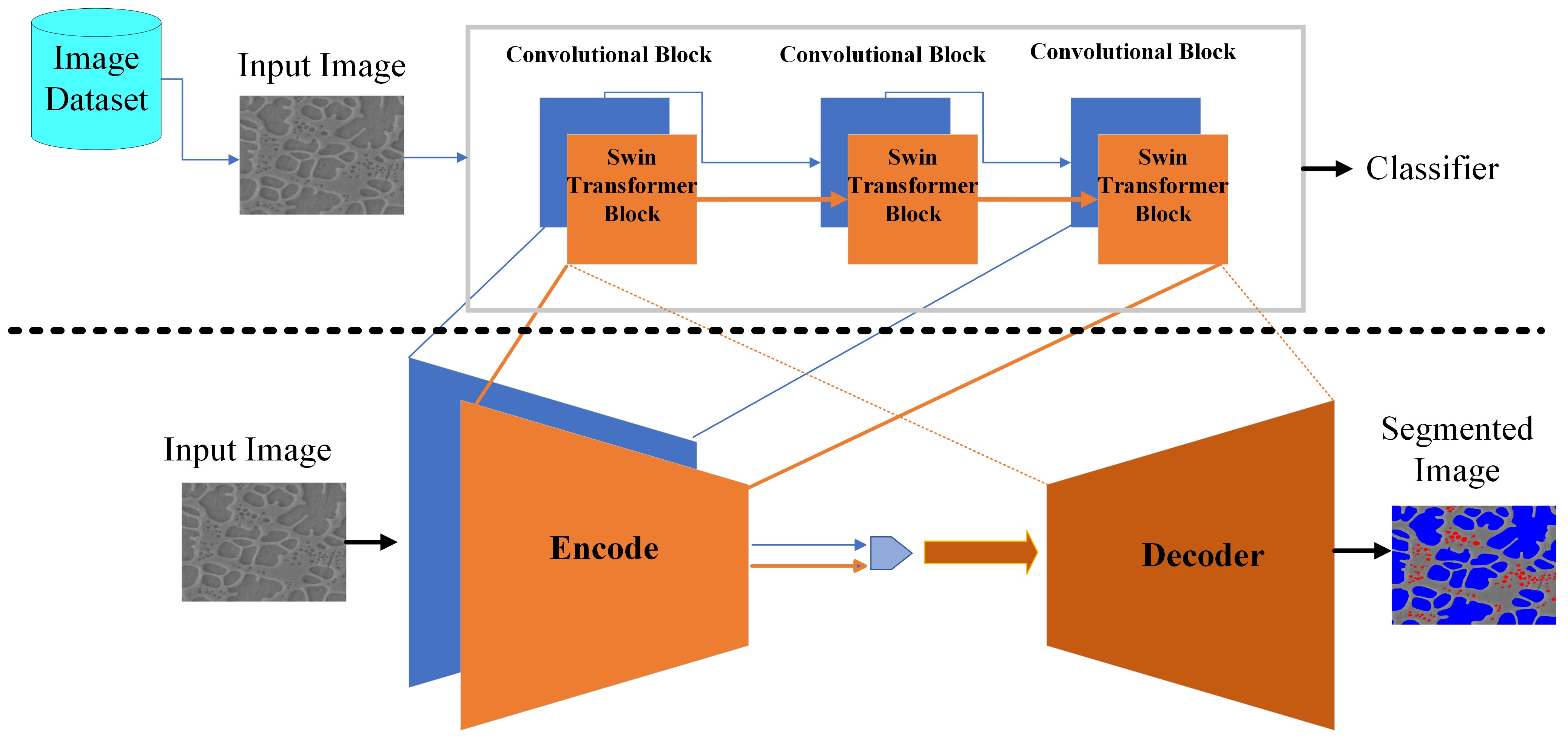}
  \caption{The encoder-decoder architecture for microstructure segmentation with transferring learning, where the CNN and Swin-T models are pre-trained on ImageNet and microscopy images. The weights of the pre-trained CNN and Swin-T models are used to initialize the encoders while the weights of the Swin-T models are used to initialize the decoders.}
\label{fig:architecture}
\end{figure}
%% TODO: change "Images Database" to "Image dataset"
%% Add a divider line between the pre-training and the segmentation parts of the image; put the phrase "pre-training" on top of the divider line

In this paper, we evaluate transfer learning using a combination of CNN and Transformer encoders for the segmentation of microscopy images. The transfer learning approach is illustrated in Figure~\ref{fig:architecture}, which contains 
an encoder-decoder architecture for image segmentation. Each encoder transforms the input image into a latent representation vector to extract semantic information. The decoder maps the extracted information back to each pixel in the input image to generate a pixel-wise classification of the image~\cite{stuckner2022microstructure,ge2020deep,jacquemet2021deep}.

%% cite paper for Swin-T (Done)
We use a popular version of Transformer -- Swin-Transformer and specifically its {\em tiny} version -- Swin-T\cite{liu2021swin} for efficiency purpose.
Our pre-training dataset contains about 50,000 microscopy images in 74 classes, which we will refer to as {\em MicroLite} dataset. 
The Swin-T model can be initialized with weights from a model pre-trained on ImageNet before fine-tuned on MicroLite.
We used CNN model pre-trained on MicroNet to initialize the weights of the blue encoder in Figure~\ref{fig:architecture} while the Swin-T model is used to initialize the weights of the orange encoder and decoder. The output of the CNN and Swin-T encoders are fused before connecting to the decoder. 
To evaluate the segmentation performance with transfer learning, we compared the IoU scores of image segmentation on 7 datasets (subsets of Super and EBC) using models pre-trained on ImageNet alone and the models pre-trained on microscopy images.  
Our results show that while the segmentation accuracy of one-shot and few-shot learning is improved, the gain is not as large as demonstrated in the NASA paper~\cite{stuckner2022microstructure}. The accuracy for out-of-distribution images remains significantly better with pre-training on microscopy images. We also compared the performance of segmentation using CNN, Swin-T, and their combination. Our results show that the combination is better than CNN alone in most cases and is better than Swin-T alone in some of the cases.

\section{Related Work}
In this section, we review prior research related to Transformers for image analysis and the Transformer-based segmentation algorithm that we have deployed and evaluated.

\subsection{Transformer}
CNN is based on the convolution operator to provide translational equivariance. However, the local receptive field of CNN limits its ability to capture long-range relation between pixels~\cite{alrfou2022computer}. Transformer has recently been used for CV tasks as a alternative to CNN.
Transformer is a type of deep neural network introduced by Vaswani {\em et al.}~\cite{vaswani2017attention}, which was successfully applied in Natural Language Processing (NLP) due to their ability to capture long-range dependencies in sequential data such as text. This approach led to significant improvements in NLP applications, such as language translation, text classification, and text generation. Compared to NLP tasks, using Transformer-based models on computer vision tasks is more challenging since images have size variations, noises, and redundant modalities.
Self-attention process is the fundamental building block of Transformer aiming to learn self-alignment that provides the ability to capture the long-range relation between image patches~\cite{alrfou2022computer}. This has led to much interest in the Transformer-based approach in CV domains~\cite{dosovitskiy2020image} such as image recognition, image segmentation~\cite{ye2019cross}, object detection~\cite{daideformable,carion2020end}, image super-resolution, and image generation~\cite{zhang2019self}.
%% In Transformer, there are typically two training stages. The first stage is pre-training, which is done on a large-scale dataset either supervised or self-supervised. Second, employing small or mid-scale datasets to fine-tune. In this stage the pre-trained weights are adjusted to the downstream tasks (e.g., image classification, image segmentation, object recognition)~\cite{dosovitskiy2020image ,alrfou2022computer}.

Dosovitskiy {\em et al.}~\cite{dosovitskiy2020image } proposed a Vision Transformer (ViT) based on a vanilla Transformer Network for NLP~\cite{vaswani2017attention }  with as few modifications as possible to capture the global context of an input image. ViT splits each image into patches and provides the Transformer with the linear embedding of each patch in order. Image patches are handled in the same manner as tokens in an NLP application. Supervised learning is used to train the model for image classification. The model is fine-tuned using downstream recognition benchmarks such as ImageNet classification after pre-training on a JFT dataset with 300 million images~\cite{sun2017revisiting }. ViT has better performance than traditional CNN and achieved 88.5\% on ImageNet classification task. However, ViT required more computational resources to train. In addition, the complexity of computing SoftMax for each self-attention block is quadratic with respect to the length of input sequence, limiting its applicability to high-resolution images~\cite{dosovitskiy2020image ,alrfou2022computer }.

To improve a Transformer model to capture local information, Liu {\em et al.}~\cite{liu2021swin } proposed a new vision Transformer called Shifted Window Transformer (Swin Transformer). This method proposed a new general-purpose backbone for image classification and recognition tasks and achieved state-of-the-art performance. The model used a shifted-window scheme to capture large variations in the scale of visual entities and high resolutions pixels in an image with linear computation complexity to input image size. In contrast, ViT~\cite{dosovitskiy2020image } model produces feature maps of a single low resolution and have quadratic computation complexity to input image size because self-attention is applied globally to all the patches. Swin Transformer achieves a good performance of 87.3\% on the ImageNet classification task, 58.7\% box average precision score on COCO detection task, and 53.5\% mIoU on the ADE20K dataset for segmentation task. 
The authors~\cite{liu2021swin} have proposed a new version of Swin Transformer V2 that would scale it up to 3 billion parameters and allow it to train with images as high quality as $1,536 \times 1,536$ pixels. 
%% TODO: check this sentence -- what does "scalable to in-capacity" mean? Swin Transformer V2 scaled up to 3 billion parameters and can be trained on images up to 1536x1536 resolution.
Swin Transformer V2~\cite{liu2022swin} modified the Swin attention module\cite{liu2021swin} for better window resolution and scale model capacity. This is done by replacing the pre-norm with post-norm configuration, using scaled cosine attention instead of dot product attention, and replacing the previous parameterized approach with a log-spaced continuous relative position bias approach~\cite{liu2022swin}.
%%to make the model salable to in-capacity and they also include a log-spaced continuous relative position bias approach that allows the model to be transferred more successfully across window resolutions~\cite{liu2022swin}.

\subsection{Segmentation}
Ronneberger et al.~\cite{ronneberger2015u,alrfou2022synergy} proposed a Fully Convolutional Network (FCN) called U-Net. U-Net is a symmetric, U-shaped, encoder-decoder neural network for semantic image segmentation. U-Net consists of a typical downsampling encoder and upsampling decoder structure and a ``skip connection'' between them. These connections copy feature maps from the encoder and concatenate them with the feature maps in the decoder.

Xie {\em et al.}~\cite{xie2021segformer } proposed a semantic segmentation framework called SegFormer that combines Transformers with a lightweight multilayer perceptron (MLP) decoder. SegFormer is based on an encoder-decoder architecture, where the encoder is a hierarchically structured Transformer that outputs multi-scale features without the need for positional encoding and the lightweight All-MLP decoder aggregates the information from different layers and combines local and global attention to produce the final semantic segmentation mask. SegFormer uses a patch size of $4 \times 4$ pixels to output a segmentation map. This approach helps improve dense prediction tasks and has resulted in impressive mIoU scores of 50.3\% on the ADE20K dataset and 84\% on the Cityscapes dataset.

Chen {\em et al.}~\cite{chen2021transunet} proposed TransUNet, a U-shaped architecture that employs a hybrid CNN-Transformer encoder followed by multiple up-sampling layers in the CNN-decoder. This method leverages both detailed high-resolution spatial information from CNN features and the global context encoded by Transformers. The TransUNet architecture includes 12 Vision Transformer (ViT~\cite{dosovitskiy2020image}) layers in the encoder, which encode tokenized image patches obtained from CNN layers. These encoded features are then upsampled using up-sampling layers in the decoder to generate the final segmentation map, with skip-connections incorporated. TransUNet achieved high performance compared with CNN-based models.

%The CNN layers first extract a feature map from the input image. Patch embedding is then applied to $1 \times 1$ patches extracted from the CNN feature map. The Transformer encodes these patches as a sequence for extracting global contexts.

Cao {\em et al.}~\cite{cao2022swin} proposed a UNet-like pure transformer model called Swin-Unet for medical image segmentation. Swin-UNet has a U-shaped encoder-decoder architecture with skip-connections for local-global semantic feature learning by feeding the tokenized image patches into the Swin-UNet model. Both encoder and decoder structures are inspired by a hierarchical Swin-Transformer~\cite{liu2021swin } with shifted windows. 

Hatamizadeh {\em et al.}~\cite{hatamizadeh2022unetr} proposed UNEt TRansformer (UNETR) model, a U-shaped encoder-decoder architecture that uses a ViT~\cite{dosovitskiy2020image}  as an encoder to capture global multi-scale information. The Transformer encoder is connected with the CNN-decoder using skip connections to compute the final semantic segmentation output. The proposed method has excellent accuracy and efficiency in various medical datasets for image segmentation tasks. 
%The input volume is divided into a sequence of uniform non-overlapping patches, which are projected into an embedding space using a linear layer. The sequence is then added with a position embedding and used as an input to a transformer model. UNETR employs the transformer as the main encoder of a segmentation network and directly connects it to the decoder via skip connections. This is in contrast to other methods that use the transformer as an attention layer within the segmentation network. UNETR also does not rely on a backbone CNN for generating the input sequences, but instead directly utilizes the tokenized patches.
Hatamizadeh {\em et al.}\cite{hatamizadeh2022swin} also proposed a novel model for medical image segmentation of brain tumors using multi-modal MRI images. This model is called Swin UNETR. It uses the Swin Transformer~\cite{liu2021swin} as encoder and connects to a CNN-based decoder via skip connections at different resolutions. In BraTS 2021 segmentation challenge, the team showed that the proposed model for multi-modal 3D brain tumor segmentation ranked among the top-performing approaches in the validation phase. 

Heidari {\em et al.}~\cite{heidari2023hiformer} proposed HiFormer that bridges a CNN and a Transformer for medical image segmentation. HiFormer efficiently uses two multi-scale feature representations and a Double-Level Fusion (DLF) module to fuse global and local features. Extensive experiments show that HiFormer outperforms other CNN-based, Transformer-based, and hybrid methods in terms of computational complexity and quality of results. HiFormer provides an end-to-end training strategy that integrates global contextual representations from Swin Transformer and local representative features from the CNN module in the encoder, followed by a decoder that outputs the final segmentation map.

Azad {\em et al.}~\cite{azad2022transdeeplab} proposed a novel approach, TransDeepLab, which is a DeepLabV3+ architecture based on pure Transformer for medical image segmentation. The proposed model uses a hierarchical Swin-Transformer with shifted windows to model the Atrous Spatial Pyramid Pooling (ASPP) module. The encoder module splits the input image into patches and applies Swin-Transformer blocks to encode local semantic and long-range contextual representation. A pyramid of Swin-Transformer blocks with varying window sizes is designed for ASPP to capture multi-scale information. The extracted multi-scale features are then fused into the decoder module using a Cross-Contextual attention mechanism. Finally, in the decoding path, the extracted multi-scale features are up-sampled and concatenated with the low-level features from the encoder to refine the feature representation.

The hybrid architectures mentioned in this section mainly focus on replacing CNN with a Transformer in the encoder, such as UNETR and Swin UNETR, or stacking CNN with a Transformer to form a new encoder in a sequential manner, such as Transnet~\cite{zhang2021transfuse,he2022swin}.

Replacing CNN with a Transformer in the encoder gives the strength of the ability to model long distance dependency in the network. However, it results in a lack of detailed texture feature extraction due to the removal of CNNs in the encoder. Stacking CNN with a Transformer to form a new encoder fails to account for the complementary relationship between the global modeling capability of self-attention and the local modeling capability of convolution. Instead, they treat the convolutional operation and self-attention as two separate and unrelated operations~\cite{wang2023cross,zhang2021transfuse}.

To overcome these drawbacks, the encoder in CS-Unet uses CNN and Transformer in parallel to obtain rich feature information from microscopy images. CNN is used to extract low-level features and Swin-T is used to extract global contextual features, which are then fused using skip connections to the decoder at different stages/layers. Moreover, to reduce feature loss in the transmission process and increase the contextual information extracted by the Swin-T encoder, the Multi-Layer Perceptron (MLP) in two successive Swin-T blocks was replaced by Residual Multi-Layer Perceptron (ResMLP).

%% Note: I have removed FAT-NET; it used a different Transformer called:  Data-Efficient Image Transformers (DeiT); we did not discuss it in the related work. 
%%Wu {\em et al.}~\cite{wu2022fat} proposed a dual encoder-based feature adaptive transformer network based on the classical encoder-decoder architecture, called FAT-Net for medical image segmentation. The encoder consists of CNN and transformer branch to effectively capture long-range dependencies and global context information. To enhance the feature fusion from these two branches, a memory-efficient decoder and a feature adaptation module had been designed.

\section{Methodology}
Our objective is to demonstrate that Transformer-based pre-training models of microscopy images are beneficial to downstream tasks such as image segmentation and that they are more robust than CNN-based pre-training models.
To this end, we completed the following tasks.
\begin{enumerate}
    \item We collected a microscopy dataset with about 50,000 images after pre-processing (MicroLite),
    \item We pre-trained Transformer encoders on MicroLite and used them to initialize the parameters of several Transformer-based segmentation algorithms (Swin-Unet, TransDeeplabv3+, and HiFormer) and a hybrid segmentation neural network based on CNN and Transformer encoders (CS-UNet),
    \item To demonstrate the advantage of CS-UNet, we compared the top performance of CNN-based segmentation algorithms~\cite{stuckner2022microstructure} with Transformer-based segmentation algorithms and CS-UNet. These algorithms are compared using the 7 test sets of the NASA team~\cite{stuckner2022microstructure}, where the CNN encoders are pre-trained on MicroNet~\cite{stuckner2022microstructure} and Transformer encoders are pre-trained on MicroLite.
    \item To evaluate the advantage of pre-training on in-domain data, we compared the top performance of CS-UNet when it is pre-trained on ImageNet and MicroLite. 
    \item To examine the detailed effects of pre-training on in-domain data, we compared the average performance of CNN-based segmentation algorithms with different pre-training settings. Similarly, we compared the average performance of Transformer-based and hybrid segmentation algorithms with different pre-training settings and Transformer architectures. 
    \item Finally, to illustrate the robustness of our hybrid strategy, we compared the performance of the three types of segmentation algorithms averaged over all configurations.  
\end{enumerate}

\subsection{Dataset Pre-processing}
The images in our MicroLite dataset are collected from multiple sources including images from different materials and compounds using various measurement techniques such as light microscopy, SEM, TEM, and X-ray. MicroLite aggregates the Aversa dataset~\cite{aversa2018first}, UltraHigh Carbon Steel Micrograph~\cite{decost2017uhcsdb }, SEM images from the Materials Data Repository, and the images from the authors of some recent publications~\cite{christiansen2019nano,mikkelsen2021scanning,salling2022individual,masubuchi2020deep,boiko2020electron,creveling2019synthetic,klinkmuller2016properties,van1978investigation}.%% {\bf [Citations]}.
%% TODO: add citations

The Aversa dataset includes over 25,000 SEM microscopy images in 10 classes, where each class consists of images in different scales (including 1, 2, 10, 20 um and 100, 200 nm) and contrast. To properly classify these images, we used a pre-trained VGG-16 model to extract feature maps from these images and used a K-means algorithm to cluster the feature maps so that images with similar feature maps are grouped in the same class. After the pre-processing step, we obtained 53 classes. The authors of Aversa dataset manually classified a small set of the images (1038) into a hierarchical dataset where the 10 classes are further divided into 27 subclasses~\cite{aversa2018first}. Our classification of these 1038 images is largely consistent with the manually assigned subclasses. Note that we have more classes since we processed the entire Aversa dataset.

In total, MicroLite includes 50,000 microscopy images labelled in 74 classes, which are obtained using the following pre-processing steps. 
\begin{enumerate}
\item Remove any artifacts such as scale bars from the images.
\item Split the images into $512 \times 512$-pixel tiles with or without overlapping depending on the size of the original images.
\item Apply data augmentation to increase the size of the dataset.
\item Aggregate the original image, the images tiles, and the augmented images to form the final dataset.
\end{enumerate}

\subsection{Pre-training}
\begin{figure}[ht]
\centering
  \includegraphics[width=150mm]{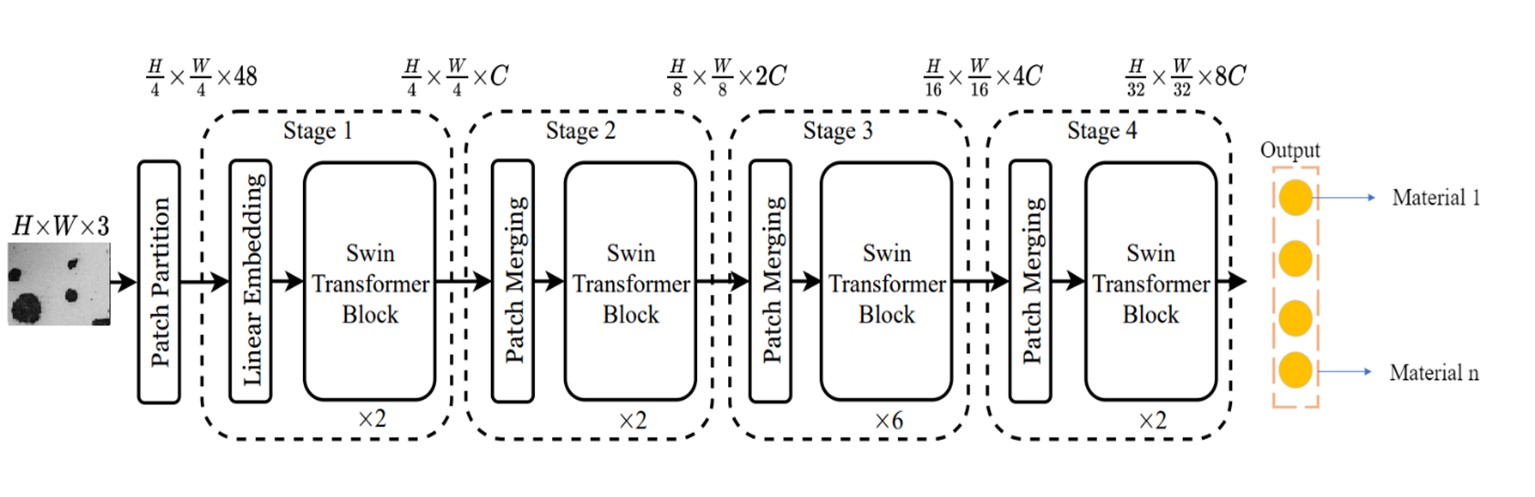}
  \caption{The original Swin-T architecture consists of four stages. Each stage contains two Swin transformer blocks except the stage three that contains six Swin transformer blocks~\cite{liu2021swin}}
\label{fig:Swin-T}
\end{figure}

We trained Swin Transformers on microscopy images to learn feature representation so that it can be transferred to tasks such as segmentation.  We evaluated two types of training. 
\begin{enumerate}
    \item Fine-tune a model pre-trained on ImageNet with MicroLite (denoted by ImageNet → MicroLite).
    \item Pre-train a model with MicroLite from scratch (denoted by MicroLite). 
\end{enumerate}
The classification tasks uses Swin-T, which is the tiny version of the Swin Transformer. Swin-T consists of two types of architectures: the original Swin-T with [2,2,6,2] transformer blocks and the intermediate network with [2,2,2,2] transformer blocks. Figure~\ref{fig:Swin-T} shows the original architecture of Swin-T. We speculate that intermediate network may be enough for microscopy analysis tasks since the earlier layers learn corner edges and shapes, the intermediate layers learn the texture or patterns, and deeper network layers in the original models learn the high-level features such as eyespots and caudal appendages.
%the position of the  animal tails or eyes.
The original and intermediate Swin-T models were pre-trained on MicroLite from scratch, where the model weights are randomly initialized. The two models were also pre-trained on ImageNet and fine-tuned on MicroLite.

The pre-training step uses an AdamW optimizer for 30 epochs with a cosine-decay learning-rate scheduler with 5 epochs of linear warm-up and batch size of 128. The initial learning rate is $10^{-3}$ and weight decay is $0.05$. The fine-tuning step also uses an AdamW optimizer for 30 epochs with a batch size of 128 but the learning rate is reduced to \(10^{-5}\) and the weight decay is reduced to \(10^{-8}\).  Models were trained until there was no improvement to the validation score using an early stopping criterion with a patience of 5 epochs. Training data had been augmented using albumentations library, which included random changes to the contrast and the brightness, vertical and horizonal flips, photometric distortions, and added noise.  

For the down-stream segmentation tasks, several models were trained for each task, which include Swin-Unet, HiFormer, and TrasDeeplapv3+. A comparative analysis was performed on the results of these models that were pre-trained with ImageNet and microscopy images.

\subsection{Combine CNN with Transformer (CS-UNet)}
\begin{figure}[ht]
\centering
  \includegraphics[width=175mm]{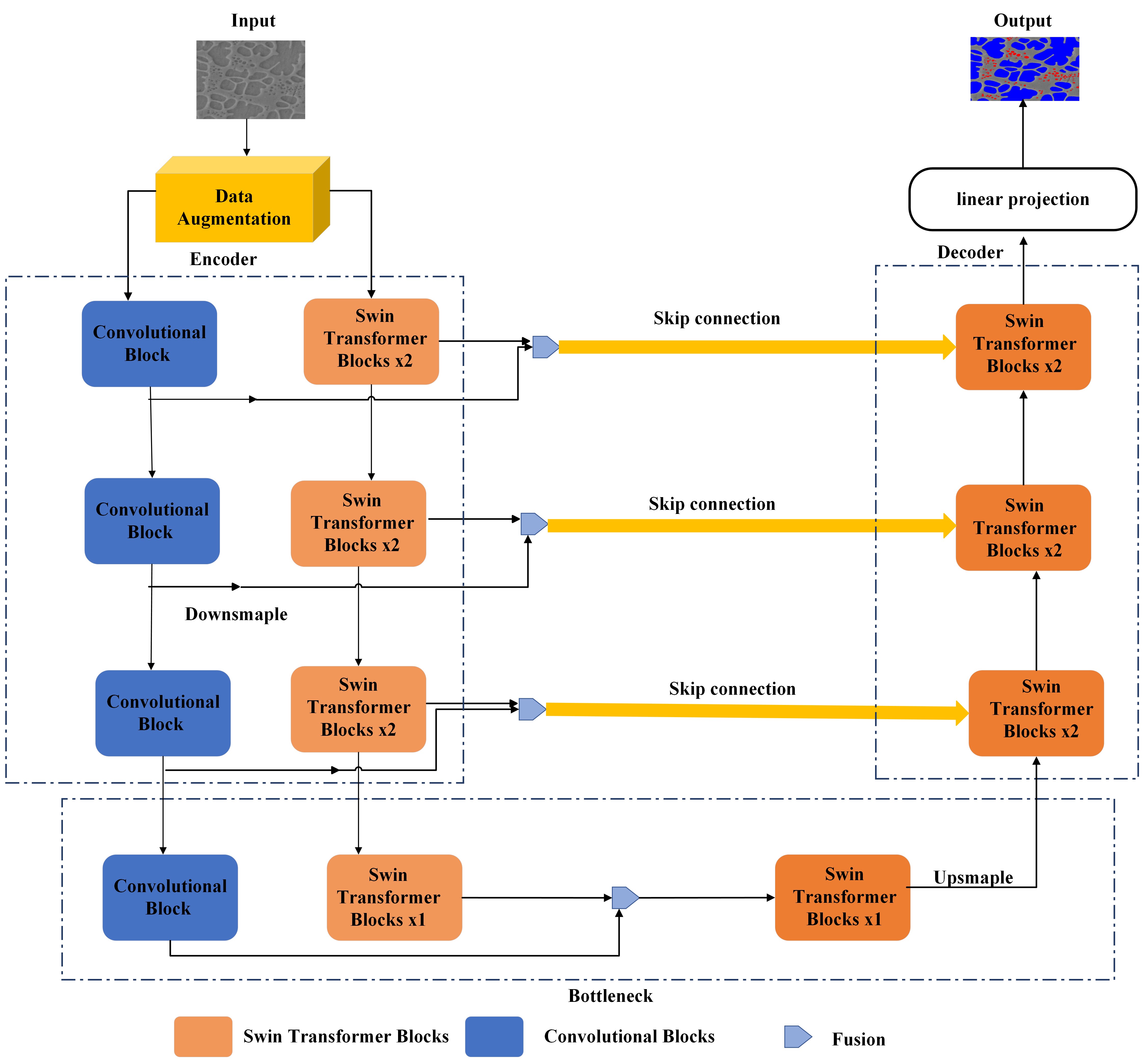}
  \caption{: Overview of the proposed CS-UNet architecture consist of encoder, bottleneck, and decoder, where the encoder part includes CNN and Swin-T.}
\label{fig:CTUNET}
\end{figure}

CNN does not capture long-range spatial relation due to its intrinsic locality. Transformer was introduced to overcome this limitation. However, Transformer has limitations in capturing low-level features. It was shown that both local and global information are essential for dense prediction tasks such as segmentation in challenging context. Some researchers introduced hybrid models that efficiently bridge a CNN and a Transformer for image segmentation. Initializing the weights of CNN and Transformer in the hybrid models will significantly improve the performance. Therefore, we introduce a hybrid UNet called CS-UNet, which is a U-shaped segmentation model that uses both CNN and Transformer. As shown in Figure~\ref{fig:CTUNET} , the method consists of encoder, bottleneck, decoder, and skip connections.

\begin{figure}[ht]
\centering
  \includegraphics[width=70mm]{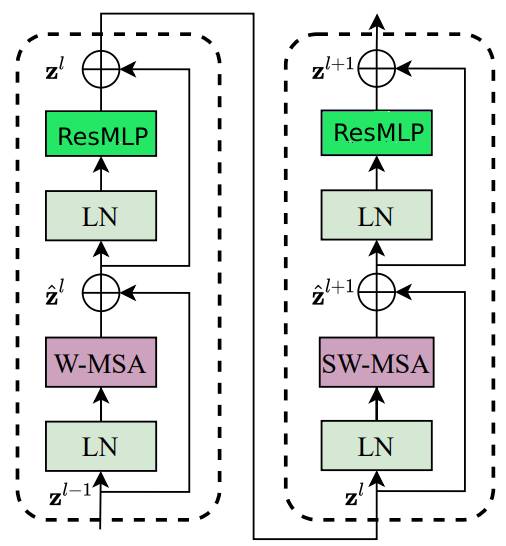}
  \caption{ResMLP used to improve Swin Transformer block~\cite{gao2022novel}.}
\label{fig:Swin_ReMLP}
\end{figure}

\begin{figure}[ht]
\centering
  \includegraphics[width=100mm]{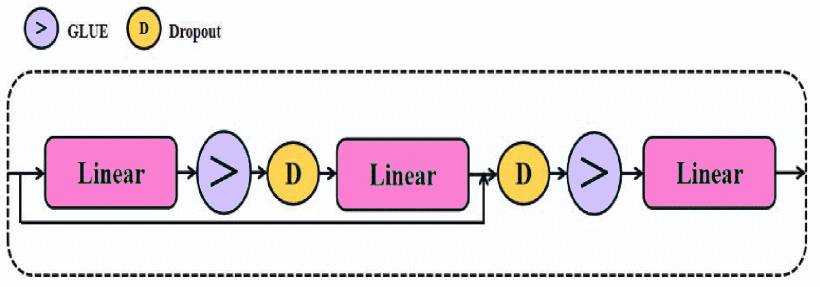}
  \caption{ResMLP module block~\cite{gao2022novel}.}
\label{fig:ReMLP_g}
\end{figure}
The encoder combines a CNN encoder and a Swin-T encoder, where CNN is used to extract low-level features and Swin-T is used to extract global contextual features. The Swin-T encoder operates on the input image divided into non-overlapping patches, applying self-attention mechanisms to capture global dependencies. The encoder captures long-range dependencies and contextual information from the entire image at different scales. Inspired by the TFCN (Transformers for Fully Convolutional dense Net)~\cite{li2022tfcns} and Lightweight Swin-Unet~\cite{gao2022novel}, the Multi-Layer Perceptron (MLP) in two successive Swin-T blocks was replaced by Residual Multi-Layer Perceptron (ResMLP).
As illustrated in Figure~\ref{fig:Swin_ReMLP}, ResMLP is used to reduce feature loss in the transmission process and to increase the contextual information extracted by the encoder. ResMLP is illustrated in Figure~\ref{fig:ReMLP_g}, which consists of two GELU~\cite{hendrycks2016gaussian} nonlinear layers, three Linear layers, and two dropout layers. 
%% TODO: ill-formed sentence
% After the second linear layer, connect with the residual and GELU layers of the original input image.
%%
The CNN encoder processes the input image in a series of convolution layers, gradually reducing the spatial dimensions while extracting hierarchical features. Along the way, the encoder captures low-level features in early layers and higher-level semantic features in deeper layers. 

To fuse the information from both encoders, the skip connections concatenate the feature maps from the CNN encoder and the Swin-T encoder with the corresponding decoder layers. To ensure compatibility between the feature dimensions of the CNN and the Swin-T encoders, it is necessary to normalize the dimensions before fusing them. This is achieved by passing the features obtained from the CNN block through a linear embedding layer, which flattens and reshapes the feature map from the shape of $(B, C, H, W)$ to $(B, C, H\times W)$, where $B$, $C$, $H$, $W$ are the batch size, the number of channels, the height, and the width of the feature map, respectively. The flattened feature map is transposed to swap the last two dimensions resulting in the shape of $(B, H\times W, C)$ and is then fused with extracted features from the Swin-T encoder. By fusing the information from different encoder pathways, the skip connections enable the decoder to benefit from both the local spatial details captured by the CNN encoder and the global context captured by the Swin-T encoder. 

The decoder is similar to that of Swin-Unet\cite{cao2022swin}, which employs the patch-expanding layer to up-sample the extracted deep features by reshaping the feature maps of adjacent dimensions to form a higher-resolution feature map, which effectively achieves a $2\times$ up-sampling. Additionally, it reduces the feature dimension to half of the original dimension. This allows the decoder to reconstruct the output with increased spatial resolution while reducing the feature dimension for efficient processing. The final patch-expanding layer further performs a $4\times$ up-sampling to restore the resolution of the feature maps to match the input resolution $(W \times H)$. Finally, a linear projection layer is applied to these up-sampled features to generate pixel-level segmentation predictions.

Different CNN families can be used in the encoder part such as EfficientNet, ResNet, MobileNet, DenseNet, VGG, and Inception. We initialize CNN weights using MicroNet and the transformer weights using MicroLite. 

\section{Results}
\label{sec:Results}
The pre-trained Swin-T models were used to classify microscopy images among 74 different classes. Swin-T models were either pre-trained on ImageNet (specifically the imageNet1K dataset) and fine-tuned on MicroLite or trained with MicroLite with randomized parameters. The training stops when the validation accuracy does not improve after 5 epochs.
The model accuracy is evaluated using the top-1 and top-5 accuracy. The top-1 accuracy measures the percentage of test samples for which the correct label is predicted while the top-5 accuracy measures the percentage of correct labeling in the top five predictions.

%% TODO: what is the stopping criteria?
As shown in Table~\ref{table:pre-trained results}, the Swin-T models, 
when trained from scratch, take longer to converge. Specifically, the original Swin-T model takes 23 epochs, while the intermediate version takes 19 epochs. In contrast, the Swin-T models pre-trained on the ImageNet and then fine-tuned on the MicroLite converge faster. The original Swin-T model takes only 13 epochs, while the intermediate version takes 12 epochs. On average, the models initialized with ImageNet weights converged about 40.16\% faster than those with the random initialization. Original Swin-T fine-tuned on MicroLite after pre-training with ImageNet achieved the top-1 accuracy of 84.63\%. Overall, the Swin-T models pre-trained on ImageNet and fine-tuned on MicroLite have higher accuracy and faster convergence.

\begin{table}[t]
\caption{Classification accuracy of pre-trained models, where the “ImageNet → MicroLite” indicate that the models were pre-trained on ImageNet and then fine-tuned on MicroLite.}
\begin{center}
\begin{tabular}{|l|l|l|l|l|}
\hline
\textbf{Swin-T architecture}&\textbf{Pre-training method}&\textbf{top-1 accuracy}&\textbf{top-5 accuracy}&\textbf{\# of epochs to converge}
\\
\hline
\multirow{2}{*}{Original}
& MicroLite & 84.23 & 95.91 & 23\\
\cline{2-5}
& ImageNet → MicroLite & {\bf 84.63} & 96.353 & 13\\
\hline
\multirow{2}{*}{Intermediate}
& MicroLite & 84.0 & 96.91 & 19\\
\cline{2-5}
& ImageNet → MicroLite & 84.45 & {\bf 97.83} & {\bf 12} \\
\hline
\end{tabular}
\end{center}
\label{table:pre-trained results}
\end{table}

\subsection{Microstructural Image Segmentation}
To evaluate how well the Swin-T models can extract feature representations, the pre-trained models were used to initialize the models for segmentation tasks. To allow comparison to the NASA study~\cite{stuckner2022microstructure}, we used the same 7 microscopy datasets derived from two materials: nickel-based super-alloys (Super) and Environmental Barrier Coatings (EBC). EBC datasets have two classes: the oxide layer and the background (non-oxide) layer, Super datasets have 3 classes: matrix, secondary, and tertiary. The number of images in each dataset split is shown in Table~\ref{table:datasets}. Super-1 and EBC-1 contain the full dataset labeled for their respective materials. Super-2 and EBC-2 have only 4 images in the training set to evaluate the performance of the models in a few shots. Super-3 and EBC-3 have only 1 image in the training set to evaluate performance during the one-shot learning. Super-4 have test images taken under different imaging and sample conditions. All segmentation models in this study were trained using PyTorch~\cite{paszke2019pytorch}.

\begin{table}[ht]
\caption{Number of Microscopy images in the training, validation, and test set for each experimental dataset}
\begin{center}
\begin{tabular}{|l|l|l|l|}
\hline
\textbf{Experiment}&\textbf{\# of training images}&\textbf{\# of validation images}&\textbf{\# of test images}\\
\hline
Super-1 & 10 & 4 & 4\\
\hline
Super-2 & 4 & 4 & 4\\
\hline
Super-3 & 1 & 4 & 4\\
\hline
Super-4 & 4 & 4 & 5\\
\hline
EBC-1 & 18 & 3 & 3\\
\hline
EBC-2 & 4 & 3 & 3\\
\hline
EBC-3 & 1 & 3 & 3\\
\hline
\end{tabular}
\end{center}
\label{table:datasets}
\end{table}

EBC and Super datasets were augmented in ways similar to the NASA study~\cite{stuckner2022microstructure}, which includes random cropping to $512\times512$ pixels, random changes to contrast, brightness, and gamma, and added blurring or sharpening. EBC dataset was horizontally flipped and Super dataset was randomly flipped vertically and horizontally and rotated. The training used an Adam optimizer with an initial learning rate of $2\times 10^{-4}$ until the validation accuracy showed no improvement for 30 epochs. Afterwards, the training continued with a learning rate of $10^{-5}$ until early stopping was triggered after an additional 30 epochs without any validation improvement. Since the datasets are imbalanced, the loss function was set by the weighted sum of Balanced Cross Entropy (BCE) and dice loss with a 70\% weighting towards BCE~\cite{stuckner2022microstructure}.  

% \usepackage{booktabs}
\begin{table}[t]
\caption{The combinations of encoders in CS-UNet, where the second column shows the pre-training models for the Swin-T encoders and the third column shows the pre-training models for the CNN encoders. The last column shows the models for both types of encoders, where `Microscopy' is either MicroNet or MicroLite.}
\begin{center}
\begin{tabular}{|l|l|l|l|}
\hline
%% \multicolumn{2}{|c|}{\textbf{CS-UNet}} \\ \hline
{\textbf{Swin-T architecture}} & {\textbf{Swin-T pre-training model}} & \textbf{CNN pre-training model} &\textbf{CS-UNet pre-training model} \\ \hline
Original & {ImageNet} & ImageNet & ImageNet\\ \hline
Original & {MicroLite} & MicroNet & Microscopy \\ \hline
Original & {ImageNet → MicroLite} & ImageNet → MicroNet & ImageNet → Microscopy \\ \hline
Intermediate & {MicroLite} & MicroNet & Microscopy\\ \hline
Intermediate & {ImageNet → MicroLite} & ImageNet → MicroNet & ImageNet → Microscopy \\ \hline
\end{tabular}
\end{center}
\label{table:combine}
\end{table}

The CS-UNet architecture is a flexible model that can be trained with different CNN families and initialized with different pre-trained models. Table~\ref{table:combine} shows the various combinations of pre-trained weights that were used to train the CS-UNet model. The second column shows the pre-trained weights that initialize the Swin-T encoder and the third column shows the pre-trained weights that initialize the CNN encoder. In the last column, we use the term {\em Microscopy} to refer to the case where CNN encoders are trained with MicroNet and Transformer encoders are trained with MicroLite.
Other combinations of pre-trained weights can also be used to train the CS-Unet model. For example, the Swin-T encoder could be initialized with the MicroLite weights and the CNN encoder could be initialized with the ImageNet→MicroNet weights.
The flexibility of the CS-UNet architecture allows researchers to experiment with different combinations of pre-trained weights to find the best combination for their specific task.

%% \FloatBarrier
%% \textcolor{blue}{The NASA paper~\cite{stuckner2022microstructure} used 35 CNN encoders to perform experiments on seven datasets. To reduce the number of experiments needed for a fair comparison, we picked 5 CNN encoders with the highest average performance in each dataset. These pre-trained encoders were included in our CS-UNet model and applied to the seven datasets. Table~\ref{table:CNNencoders} in Appendix~\ref{app:cnn} shows the 19 encoders that have top-5 accuracy for at least 1 experiment on the 7 datasets.}

\begin{table}[ht]
\caption{Comparison of the top performance (IoU) of UNet++/UNet pre-trained on MicroNet \cite{stuckner2022microstructure}, Transformer segmentation models pre-trained on MicroLite, and CS-UNet pre-trained on MicroNet and MicroLite.}
\begin{center}

\begin{tabular}{@{}|l|c|c|c|@{}}
\hline
Dataset & \multicolumn{1}{l|}{UNet++/UNet + MicroNet} & \multicolumn{1}{l|}{Transformer + MicroLite} & \multicolumn{1}{l|}{CS-UNet + MicroNet + MicroLite} \\ \hline
Super-1 & %% 95.95\% 
          96.4\% & 95.72\% & \textbf{96.43\%} \\ \hline
Super-2 & %% 95.50\% 
          94.2\% & 95.16\% & \textbf{96.06\%} \\ \hline
Super-3 & %% 92.5\% 
          93.0\% & 92.23\% & \textbf{93.5\%} \\ \hline
Super-4 & %% 78.95\% 
          78.5\% & 78.91\% & \textbf{82.13\%} \\ \hline
EBC-1 & %% 96.67\% 
          97.6\% & 96.59\% & \textbf{97.66\%} \\ \hline
EBC-2 & %% 90.97\% 
          \textbf{93.3}\% & 91.11\% & 92.82\% \\ \hline
EBC-3 & %% 52.84\% 
          65.9\% & \textbf{82.13\%} & 70.46\% \\ \hline
\end{tabular}
\end{center}
\label{table:compare}
\end{table}

Table~\ref{table:compare} compares the top performance of UNet++/UNet pre-trained on MicroNet (the numbers were reported in Figure 3--5 of~\cite{stuckner2022microstructure}), Transformer models (including Swin-UNet, TransDeepLabV3+, and HiFormer) pre-trained on MicroLite, and CS-UNet pre-trained on MicroNet and MicroLite. The highest accuracy for each experiment is shown in bold font. 
CS-UNet has the best performance in most experiments except EBC-2 and EBC-3. 
For experiments with ample training data such as Super-1 and EBC-1, the difference between UNet++/UNet, Transformer, and CS-UNet is small. 
For few-shot learning experiments such as Super-2 and EBC-2, the accuracy gain of CS-UNet is modest.
For one-shot learning experiments, the result is mixed, where CS-UNet has modest improvement in Super-3 while significant gain in EBC-3.
For out of context learning, CS-UNet shows significant improvement over UNet or Transformer. 

Overall, the results of the table suggest that CS-UNet is a promising approach for image segmentation tasks. CS-UNet is similar or significantly better than UNet++/UNet in all experiments and it is better than Transformers for most experiments. 
Note that MicroLite is about half the size of MicroNet. Despite this, 
Transformer + MicroLite has comparable or better performance than that of UNet++/UNet + MicroNet. 
Appendix~\ref{sec:Trans} shows the configurations of the best performing Transformer models shown in Table~\ref{table:compare}. The configurations of the best performing CS-UNet models are shown in the next section, where
we compare the performance of CS-UNet when it is pre-trained on microscopy images and on ImageNet. 

%% where Appendix~\ref{sec:unets} shows the best encoders for the UNet for each dataset that pre-trained on both ImageNet and MicroNet.}
%% The results also shows that training the CNNs and Swin-T encoders with large microscopy datasets achieved better performance, especially when we a few-shot and one-shot training image. These results also suggest that the MicroNet and MicroLite models are able to generalize well when the test set has different conditions compared to the training set, even when they are only trained on a small number of images

\subsection{Nickel-based Super-alloy (Super) Segmentation}
\begin{figure}[!ht]
\centering
  \includegraphics[scale=0.50]{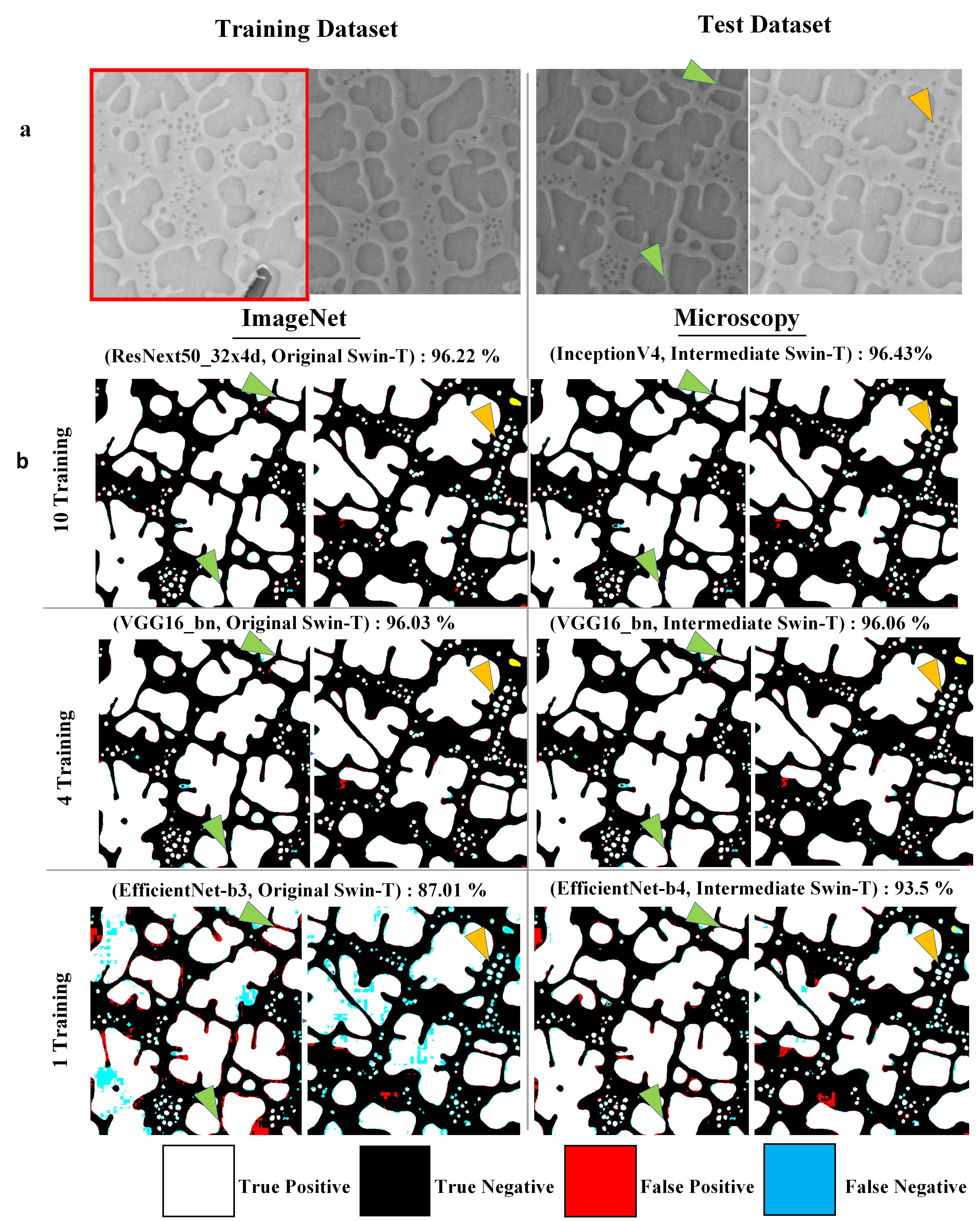}
  \caption{(a) are examples of the training and test images of the Super datasets~\cite{stuckner2022microstructure}, where the training image of Super-3 is outlined in red. (b) are the best segmentation masks of the test images when CS-UNet is trained with Super-1, Super-2, and Super-3 dataset, where the left is pre-trained with ImageNet and the right is pre-trained with microscopy images. 
  The CNN/Transformer encoders and the IoU score of the best model are above the segmentation mask of each experiment.
  The green triangle points to the area where the ImageNet (but not the microscopy) model incorrectly segmented the secondary precipitates. The yellow triangle points to the area where the ImageNet (but not the microscopy) model incorrectly segmented the tertiary precipitates. 
  }
\label{fig:Super1-3}
\end{figure}

\begin{figure}[ht]
\centering
  \includegraphics[scale=0.50]{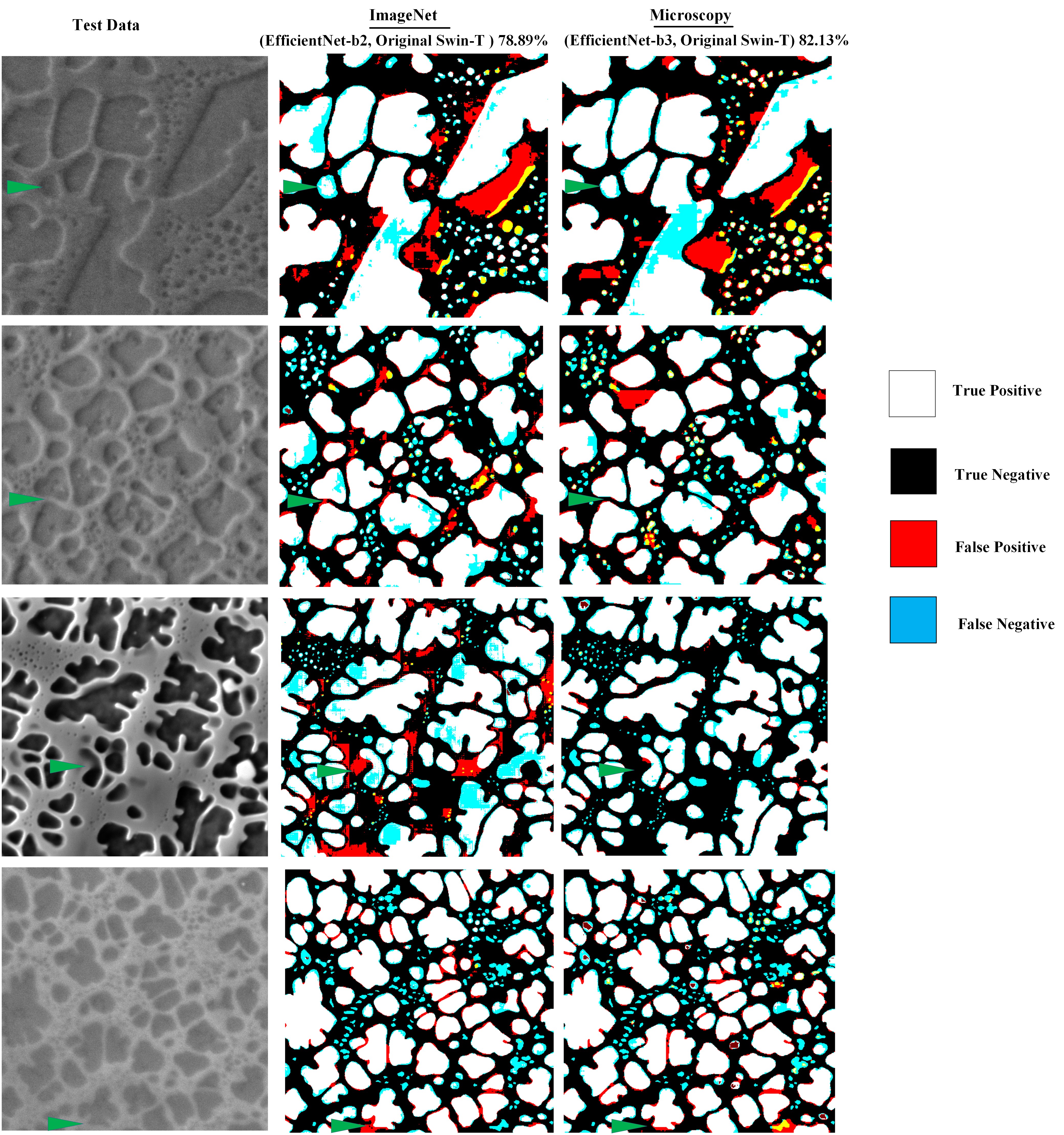}
  \caption{The test images of Super-4 have different imaging conditions than those of the training images. The left column shows the test images of Super-4~\cite{stuckner2022microstructure}. The accuracy mask colors are the same as those in Figure~\ref{fig:Super1-3}. The middle column shows the IoU accuracy masks for the best ImageNet model. The right column shows the same for the best microscopy model. Each row shows the test image and accuracy masks of the same image. The green arrow shows an example where the model was over-segmented and where the microscopy model accurately segmented the secondary precipitate. The yellow color indicates incorrect identification of a tertiary precipitate as a secondary precipitate. The maroon color indicates where the model improperly identified a secondary precipitate as a tertiary precipitate. The cyan color indicates where the model over-segmented the secondary precipitates.}
\label{fig:Super4}
\end{figure}

Figure~\ref{fig:Super1-3} and Figure~\ref{fig:Super4} compare the best performance of CS-UNet on Super datasets when it is pre-trained on the microscopy images and ImageNet. 
The IoU scores for Super-1 and Super-2 are similar for both pre-training models while the IoU score for Super-3 is significantly higher for microscopy model at 93.5\% in comparison to the 87.01\% of the ImageNet model. 
This result differs from that of the NASA paper, where the performance of Super-2 was also increased. It appears that with the enhanced ability of CS-UNet, the benefit of in-domain dataset is more significant in one-shot learning than few-shot learning.  
%%
%% Pre-training on Microscopy resulted in a substantial improvement in accuracy for one-shot learning on the Super-3 dataset. The best performance of CS-UNet pre-trained on Microscopy and ImageNet for each Super-1 to Super-3 dataset are shown in Figure~\ref{fig:Super1-3}.
%%
%% The left side of Figure~\ref{fig:Super1-3}-a hows two example training images included in the Super datasets, where the image outlined in red is the sole image of Super-3~\cite{stuckner2022microstructure}. The right side of Figure~\ref{fig:Super1-3}-a shows two test images included in all 3 Super datasets. 
%% Figure~\ref{fig:Super1-3}-b shows the best segmentation results on the test images of Figure~\ref{fig:Super1-3}-a using models trained on each Super dataset with initial parameters pre-trained on (MicroLite + MicroNet) and ImageNet.
%% For Super-1, both Microscopy and ImageNet models accurately segmented the secondary and tertiary precipitates with IoU of 96.43\% and 96.32\%, respectively. For Super-2, there is a slight reduction in the accuracy for both models, with Microscopy at 96.06\% and ImageNet at 96.03\%. 
%%
%% For Super-3, microscopy outperforms ImageNet with an accuracy of 93.5\% IoU, compared to 87.01\% IoU for ImageNet. 
The ImageNet model failed to identify many of the tertiary precipitates in the dark contrast images, as indicated by the orange triangles. The ImageNet model also over-segmented and combined some of the secondary precipitates, as indicated by the green arrows.

%% \FloatBarrier

%% The orange and green triangles also show that reducing the number of training images affects the segmentation results. For example, the accuracy score for ImageNet on Super-1 is 96.32\% when the number of training images is 10, but it decreases to 96.03\% when the number of training images is 4. The decrease in accuracy score is more pronounced for datasets with one shot. For example, the accuracy score for ImageNet on Super-2 is 96.03\% when the number of training images is 4, but it decreases to 87.01\% when the number of training images is 1.
%% The accuracy score (IoU) for Microscopy is more stable as the number of training images decreases. For example, the accuracy score for Microscopy on Super-1 is 96.43\% when the number of training images is 10, and it remains at 96.06\% when the number of training images is 4.
%% For full and few shot training images, the segmentation outputs of both models can help identify the precipitate boundaries in images of both bright and dark contrast. The tertiary precipitates were recognized, and the secondary precipitates were appropriately separated.

%% For Super-3, the Microscopy model achieved a remarkable accuracy of 93.0\% IoU when trained with only one image, which is almost as accurate as training on the entire dataset. The accuracy score for Microscopy is also higher than the accuracy score for ImageNet on all Super datasets.

For Super-4, which contains images with different imaging conditions, microscopy model improves the performance of ImageNet model from 78.89\% to 82.13\%, which is consistent with the findings of the NASA paper. As shown in Figure~\ref{fig:Super4}, the test images of Super-4 comes from a different image distribution than the training images (Figure~\ref{fig:Super1-3}). The first row shows a micrograph that comes from a different alloy. Rows (2 and 3) show micrographs with different etching conditions, and the last row shows micrograph with poor imaging~\cite{stuckner2022microstructure}.
The microscopy model is more accurate and less over-segmented for separating the secondary precipitates than the ImageNet model, where the differences are marked with green arrow. The microscopy model performs better for the secondary and the tertiary precipitate of the segmented images. 

 \FloatBarrier

\subsection{Environmental Barrier Coating (EBC) Segmentation}
%% The top model pre-trained on microscopy images achieved significant improvement in the one-shot learning case on the EBC-3 dataset, compared to the top ImageNet model. 
%% As shown in Table~\ref{table:encoder-performance}, ImageNet models performed slightly better on average for EBC-1 and EBC-3, while pre-training on ImageNet-then-Microscopy gave the best average performance on EBC-2.

\begin{figure}[ht]
\centering
  \includegraphics[width=175mm]{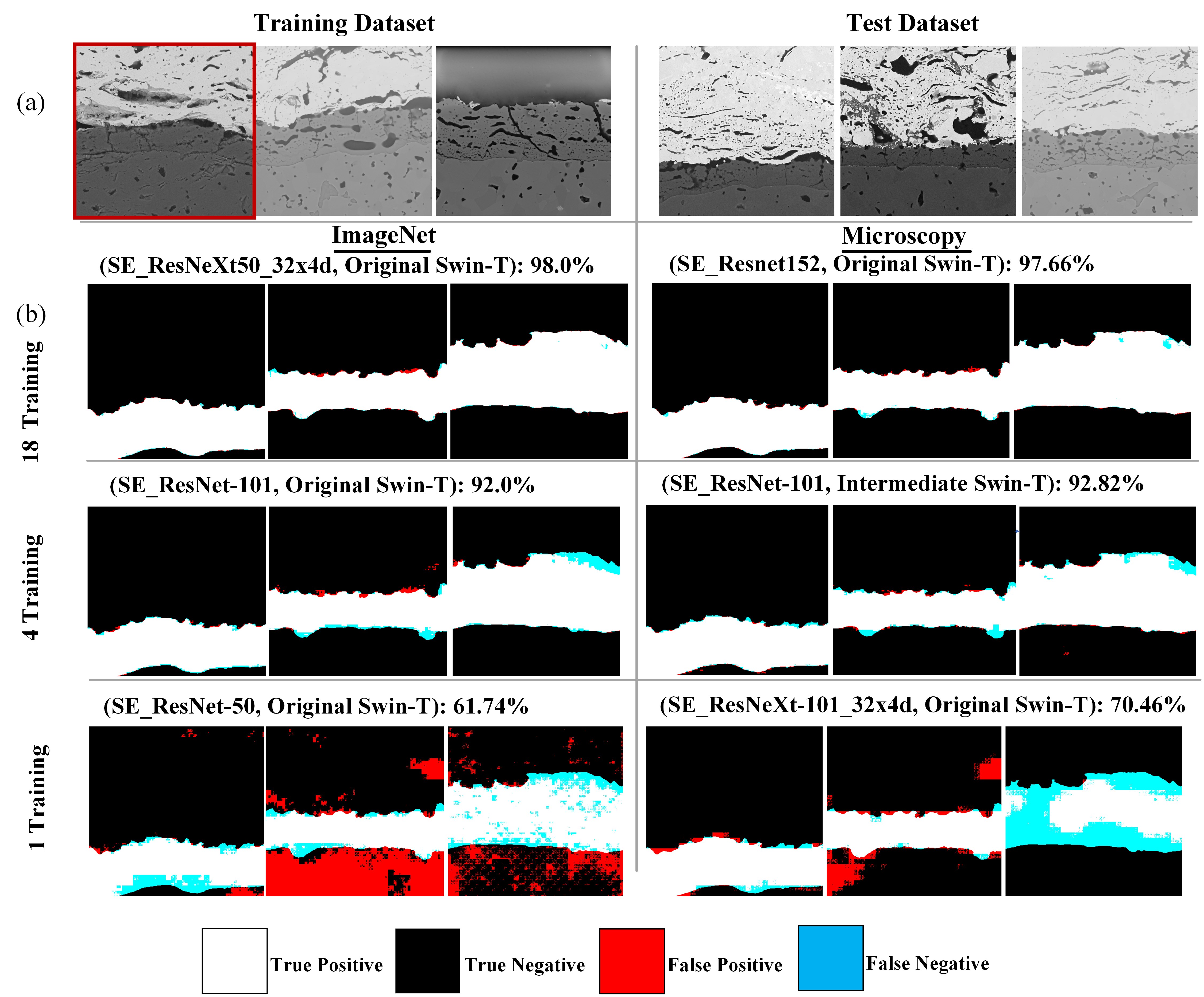}
  \caption{(a) shows examples of the train and test images in the EBC datasets, where the left column represents the training set and the right column represents the test set. The single training image for EBC-3 is outlined in red. (b) shows the segmentation results for the top ImageNet and microscopy models for each EBC dataset. The CNN/Transformer encoders and the IoU score of the best model is above the segmentation mask of each experiment.}
\label{fig:EBCs}
\end{figure}

As shown in Figure~\ref{fig:EBCs}, 
the results of EBC datasets are consistent with the findings of the NASA paper, where the IoU scores of EBC-1 and EBC-2 are similar for both the microscopy and the ImageNet models.
%% For EBC-1, both (MicroLite + MicroNet) and ImageNet models performed well and accurately segmented the thermally grown oxide layer with IoU of 97.66\%  and 98.0\%  for microscopy and ImageNet, respectively. 
%% For EBC-2, which contains only four training microscopy images, the accuracy was reduced for both microscopy of 92.0\% and ImageNet of 92.82\%. 
The segmentation outputs of EBC-1 and EBC-2 for both models are able to distinguish between the substrate and the thermally grown oxide layer and they are useful for oxide thickness measurements after simple morphological operations.
%%Note: EBC-3 result using Microscopy was 72.0 for CS-Unet(ResNet-18), ResNet-18 is not of the Top 5 encoders. For EBC-2 using microscopy was 94.2% for CS-Unet (Efficeintnet-b3), Efficeintnet-b3 is not in the top 5 Encoders.

EBC-3 has one training image (outlined in red in Figure~\ref{fig:EBCs}). The IoU score of the best microscopy model is 70.46\% much higher than the IoU of 61.74\% of the best ImageNet model. 
This also confirms the result of the NASA paper though our improvement is much larger.
The ImageNet model is not able to distinguish between the substrate and the thermally grown oxide layer, which made it impossible to accurately measure oxide thickness.

 \FloatBarrier

\section{Discussion}
We have shown that CNN and Transformer encoders in CS-UNet provide better segmentation performance than CNN encoders alone in UNet. We have also shown that pre-training on microscopy images provide better performance for CS-UNet than pre-training on ImageNet, though the degree of improvement differs from that observed on UNet. 
Nevertheless, these comparisons are based on best performing models. Depending on the choice of CNN encoders, Transformer architectures, and pre-training models, the segmentation performance can vary significantly.

In this section, we compare the impact of pre-training on the average performance of CNN-based, Transformer-based, and hybrid segmentation algorithms. After that, we compare the average performance of the three types of segmentation algorithms. Our results indicate that pre-training on microscopy images generally has positive impacts to the performance. Our hybrid algorithm CS-UNet outperforms UNet in all experiments while it has similar or better performance than Transformer-based algorithms.

%% This section compares image segmentation performance on  CNN, Transformers, and CS-Unet models under the IoU metric. The NASA team\cite{stuckner2022microstructure} trained CNN models with about 110,000 microscopy images in MicroNet. We trained Swin-T models with 50,000 microscopy images.

%% After exploring the performance of CS-UNet, a hybrid method that combines the strengths of CNNs and Transformers for image segmentation, we evaluated the performance of pre-trained Swin-T in MicroLite using different  Transformer vision segmentation methods. We compared the results with CS-UNet and the MicroNet pre-training model in the CNN method.

%% We aimed to determine whether the initialization weights from the pre-trained CS-UNet model offer better performance than those from pre-trained CNN and Transformer models. We trained  the transformer method, including Swin-Unet and TransDeepLabv3+, and a CNN model, including UNet using different pre-trained models as mentioned in section~\ref{sec:Results}.

\subsection{CNN-based Image Segmentation}

\begin{table}[ht]
\caption{The average performance of {\bf UNet} when it is initialized with different pre-training weights. Each entry shows the mean and standard deviation of IoU value for a pre-training model. The highest score for each test is in bold.}
\begin{center}
\begin{tabular}{@{}|l|l|l|l|@{}}
\hline
 \textbf{Dataset} &  \textbf{ImageNet} & \textbf{MicroNet} &  \textbf{ImageNet→MicroNet} \\ \hline
Super-1 & \textbf{96.09\%±0.20\%} & 95.99\%±0.16\% & 95.92\%±0.19\% \\ \hline
Super-2 & 95.08\%±1.27\% & 95.28\%±0.26\% & \textbf{95.38±1.03\%} \\ \hline
Super-3 & 63.30\%±4.55\% & 74.61\%±17.21\% & \textbf{78.69±11.24\%} \\ \hline
Super-4 & 71.46\%±6.88\% & 75.1\%±9.0\% & \textbf{77.78\%±0.17} \\ \hline
EBC-1 & 95.18\%±0.82\% & 94.65\%±1.44\% & \textbf{95.69\%±1.03\%} \\ \hline
EBC-2 & 80.74\%±10.76\% & \textbf{87.06\%±1.55\%} & 86.0\%±5.17\% \\ \hline
EBC-3 & 39.36\%±7.81\% & 41.95\%±4.89\% & \textbf{46.44\%±4.35\%} \\ \hline
\end{tabular}
\end{center}
\label{table:Unet}
\end{table}

We first examine the performance of UNet~\cite{ronneberger2015u} with 3 types of encoder pre-training. 
We include this result since the configurations of CS-UNet only used 
19 of the 35 CNN encoders in the NASA paper~\cite{stuckner2022microstructure}. 
As shown in Appendix~\ref{app:cnn}, the 19 encoders have top-5 accuracy in at least one of the segmentation tasks. 
This selection reduces the number of experiments needed for a fair comparison, 
%% These pre-trained encoders were included in our CS-UNet model and applied to the 7 segmentation tasks.

The average performance of UNet when it is pre-trained with ImageNet or MicroNet is in Table~\ref{table:Unet}, which indicates that ImageNet→MicroNet model (i.e. CNN encoders initialized with ImageNet model and fine-tuned with MicroNet) achieves the best outcome in majority of the cases. The configurations of the top-performing CNN encoders are shown in Appendix~\ref{app:Unet}, which also indicates that pre-training on MicroNet provides better outcome in most of the cases.

%% The ImageNet pre-training model performs well on most datasets, with slightly better accuracy for Super-1. However, it achieves lower segmentation results on Super-3 and EBC-3 compared to the other pre-training models. The MicroNet pre-training method shows competitive performance across the datasets, consistently achieving high segmentation results. Especially for EBC-2, MicroNet outperforms ImageNet and is slightly better than ImageNet→MicroNet. The ImageNet→ MicroNet pre-training model demonstrates improved performance compared to both ImageNet and MicroNet pre-training methods, especially on Super-3 and EBC-3 (one-shot learning), where it outperforms the other models.

%% We found that training the UNet model using ImageNet→MicroNet resulted in better performance, which is consistent with the findings of the NASA team in their paper~\cite{stuckner2022microstructure}. On average, MicroNet and ImageNet→MicroNet had better performance for few-shot and one-shot learning.

Unsurprisingly, our results are largely consistent with that of the NASA paper~\cite{stuckner2022microstructure}, since we used their pre-training models on MicroNet. Specifically, pre-training with MicroNet improves the performance of one-shot and out-of-distribution learning. Since we picked CNN encoders that have top-5 performance in at least one experiment, the IoU scores are higher than the average scores shown in the NASA paper. In fact, the performances on Super-2 (few-shot learning) are basically the same with different pre-training models. 

\subsection{Transformer-based Image Segmentation}

%% To evaluate the feature extraction capabilities of Swin-T models, we fine-tuned them on the pre-training MicroLite for segmentation tasks. We trained different transformer vision segmentation methods on the same seven datasets with different pre-trained weights.

\begin{table}[ht]
\caption{Average performance of {\bf Transformer}-based segmentation algorithms when initialized with different pre-training weights.
Each entry shows the mean and standard deviation of IoU value for a pre-training model. The highest score for each dataset is in bold.}
\label{table:Transformer-performance}
\begin{tabular}{@{}|l|lll|ll|@{}}
\hline
\multicolumn{1}{|c|}{\textbf{Swin-T arch.}} & \multicolumn{3}{c|}{\textbf{Original}} & \multicolumn{2}{c|}{\textbf{Intermediate}} \\ \hline
\textit{\textbf{Dataset}} & \multicolumn{1}{l|}{\textit{\textbf{ImageNet}}} & \multicolumn{1}{l|}{\textit{\textbf{MicroLite}}} & \textit{\textbf{ImageNet→MicroLite}} & \multicolumn{1}{l|}{\textit{\textbf{MicroLite}}} & \textit{\textbf{ImageNet→MicroLite}} \\ \hline
Super-1 & \multicolumn{1}{l|}{94.94\%±0.31\%} & \multicolumn{1}{l|}{\textbf{95.0\%±0.43\%}} & 94.89\%±0.45\% & \multicolumn{1}{l|}{94.72\%±0.97\%} & 94.41\%±0.51\% \\ \hline
Super-2 & \multicolumn{1}{l|}{93.26\%±0.76\%} & \multicolumn{1}{l|}{93.83\%±0.62\%} & 93.55\%±0.30\% & \multicolumn{1}{l|}{\textbf{94.03\%±0.83\%}} & 93.43\%±0.97\% \\ \hline
Super-3 & \multicolumn{1}{l|}{76.24\%±7.06\%} & \multicolumn{1}{l|}{\textbf{79.53\%±13.69\%}} & 70.65\%±10.74\% & \multicolumn{1}{l|}{78.74\%±17.08\%} & 69.31\%±6.44\% \\ \hline
Super-4 & \multicolumn{1}{l|}{73.04\%±1.64\%} & \multicolumn{1}{l|}{\textbf{73.18\%±1.81\%}} & 72.89\%±2.93\% & \multicolumn{1}{l|}{72.10\%±2.39\%} & 71.23\%±2.32\% \\ \hline
ECB-1 & \multicolumn{1}{l|}{91.73\%±3.40\%} & \multicolumn{1}{l|}{94.67\%±1.59\%} & 91.44\%±2.87\% & \multicolumn{1}{l|}{\textbf{94.95\%±1.41\%}} & 90.56\%±3.01\% \\ \hline
ECB-2 & \multicolumn{1}{l|}{82.47\%±5.90\%} & \multicolumn{1}{l|}{86.02\%±4.98\%} & 83.20\%±3.34\% & \multicolumn{1}{l|}{\textbf{86.67\%±2.61\%}} & 83.07\%±4.22\% \\ \hline
ECB-3 & \multicolumn{1}{l|}{52.21\%±6.76\%} & \multicolumn{1}{l|}{\textbf{65.90\%±13.12\%}} & 55.15\%±10.42\% & \multicolumn{1}{l|}{56.91\%±5.24\%} & 49.03\%±4.67\% \\ \hline
\end{tabular}
\end{table}

Next, we show in Table~\ref{table:Transformer-performance}
the average performance of Transformer-based segmentation algorithms (Swin-UNet, HiFormer, and TransDeepLabv3+) using different configurations of pre-training and Swin-T architectures.
We compared the algorithms by using the original or the intermediate Swin-T architecture and by initializing their weights with ImageNet or microscopy pre-training models. 
Our results indicate that the algorithms perform well with the MicroLite pre-training model and that the original Swin-T architecture is slightly better with 1-shot learning and out-of-distribution learning.
Overall, pre-training with microscopy images provided better results for Transformer-based segmentation algorithms than pre-training on natural images.

\subsection{Hybrid Image Segmentation}

%% The pre-trained weights have an impact on the performance of CS-UNet. The models that are pre-trained on MicroLite generally outperform the models pre-trained on ImageNet when there are limited training data. Table~\ref{table:encoder-performance} shows the performance of CS-UNet for each experiment, averaged over all combinations of CNN and Swin-T encoders and initialized with different pre-training weights. The results show that, for few-shot or one-shot learning as in Super-2, Super-3, and ECB-2, pre-training with ImageNet-then-MicroLite was slightly better than pre-training with MicroLite or ImageNet alone. When averaged over all experiments, the combination of pre-training with ImageNet-then-MicroLite and intermediate Swin-T architecture produced the best performance. 

% \usepackage{booktabs}
\begin{table}[ht]
\caption{Average performance of {\bf CS-UNet} when initialized with different pre-training weights for each experiment. Each entry in the table shows the mean value and standard deviation of the evaluation IoU metric for a particular pre-training model. The highest accuracy score for each dataset is shown in bold.}
\label{table:encoder-performance}
\begin{tabular}{@{}|l|lll|ll|@{}}
\hline
\multicolumn{1}{|c|}{\textbf{Swin-T arch.}} & \multicolumn{3}{c|}{\textbf{Original}} & \multicolumn{2}{c|}{\textbf{Intermediate}} \\ \hline
\textit{\textbf{Dataset}} & \multicolumn{1}{l|}{\textit{\textbf{ImageNet}}} & \multicolumn{1}{l|}{\textit{\textbf{Microscopy}}} & \textit{\textbf{ImageNet → Microscopy}} & \multicolumn{1}{l|}{\textit{\textbf{Microscopy}}} & \textit{\textbf{ImageNet → Microscopy}} \\ \hline
Super-1 & \multicolumn{1}{l|}{96.11\%±0.11\%} & \multicolumn{1}{l|}{96.19\%±0.13\%} & 96.16\%±0.14\% & \multicolumn{1}{l|}{\textbf{96.22\%±0.15\%}} & 96.02\%±0.34\% \\ \hline
Super-2 & \multicolumn{1}{l|}{95.63\%±0.23\%} & \multicolumn{1}{l|}{95.63\%±0.44\%} & 95.78\%±0.14\% & \multicolumn{1}{l|}{95.67\%±0.46\%} & \textbf{95.86\%±0.11\%} \\ \hline
Super-3 & \multicolumn{1}{l|}{78.64\%±9.3\%} & \multicolumn{1}{l|}{76.59\%±15.78\%} & 78.01\%±13.32\% & \multicolumn{1}{l|}{78.18\%±12.13\%} & \textbf{80.68\%±12.60\%} \\ \hline
Super-4 & \multicolumn{1}{l|}{73.74\%±3.74\%} & \multicolumn{1}{l|}{\textbf{77.25\%±3.31\%}} & 72.87\%±6.26\% & \multicolumn{1}{l|}{76.65\%±2.53\%} & 75.07\%±1.10\% \\ \hline
ECB-1 & \multicolumn{1}{l|}{\textbf{97.09\%±0.86\%}} & \multicolumn{1}{l|}{95.48\%±1.04\%} & 96.4\%±0.79\% & \multicolumn{1}{l|}{94.72\%±1.37\%} & 96.21\%±0.78\% \\ \hline
ECB-2 & \multicolumn{1}{l|}{83.57\%±7.71\%} & \multicolumn{1}{l|}{86.12\%±1.76\%} & \textbf{88.58\%±1.46\%} & \multicolumn{1}{l|}{86.41\%±1.65\%} & 88.08\%±2.98\% \\ \hline
ECB-3 & \multicolumn{1}{l|}{45.88\%±10.12\%} & \multicolumn{1}{l|}{44.7\%±8.57\%} & {46.08\%±13.92\%} & \multicolumn{1}{l|}{\textbf{46.35\%±9.91\%}} & 45.16\%±10.58\% \\ \hline
\textbf{mean IoU} & \multicolumn{1}{l|}{81.5} & \multicolumn{1}{l|}{81.71} &81.98 & \multicolumn{1}{l|}{82.03} &\bf{82.44} \\ \hline
\end{tabular}
\end{table}

We also compare the performance of our hybrid segmentation algorithm CS-UNet in Table~\ref{table:encoder-performance} when it uses the original or the intermediate Swin-T architecture and when it is initialized with weights from ImageNet or microscopy models.
Since CS-UNet uses both CNN and Transformer encoders, the results are mixed where pre-training with microscopy images does not provide better performance in all cases. The weaker performance of CNN encoders reduced the advantage of Transformer encoders when they are pre-trained on microscopy images.  
When we consider the mean IoU scores across all experiments, however, pre-training with microscopy images still offers some benefits.

\subsection{Comparison of Segmentation Networks}
%% This section aims to compare the performance of CNN-based UNet, Transformer-based UNet, and CS-UNet across the seven datasets, shedding light on their efficacy for different segmentation challenges.

%% We evaluate the performance of four methods: UNet, CS-UNet, Swin-Unet, HiFormer, and TransDeepLabv3+. Each method is trained and tested on seven datasets: Super-1, Super-2, Super-3, Super-4, EBC-1, EBC-2, and EBC-3. The IoU evaluation metric is utilized to measure the accuracy of segmentation results. Mean values and standard deviations are reported to provide a comprehensive analysis of the methods' performance. CS-UNet consistently outperforms UNet across all datasets as shown in Tabel~\ref{table:model-performance}. This suggests that the hybrid CNN and Swin Transformer architecture of CS-Unet is more effective for image segmentation tasks than the pure CNN architecture of Unet. CS-Unet is also competitive with Swin-Unet, HiFormer, and TransDeepLabv3+ on some datasets. This suggests that CS-Unet is a promising approach for image segmentation tasks, and that it can achieve state-of-the-art performance on some datasets.

\begin{table}[ht]
\caption{ The average performance of CNN, Transformer, and CS-UNet on all datasets. Each entry in the table shows the mean value and standard deviation of the IoU evaluation metric for a particular method.}
\begin{center}
\begin{tabular}{|l|l|l|l|l|l|}
\hline
\textbf{Dataset}&\textbf{UNet~\cite{ronneberger2015u}} &\textbf{CS-UNet}&\textbf{Swin-Unet~\cite{cao2022swin}}&\textbf{TransDeepLabV3+~\cite{azad2022transdeeplab}} &\textbf{HiFormer~\cite{heidari2023hiformer}}\\
\hline
Super-1 & 95.98\% $\pm$ 0.20\% &{\bf 96.14}\% $\pm $ 0.21\% & 95.30\%$ \pm0.34 $ \% & 94.91\% $\pm$ 0.47\% & 94.30\% $\pm $ 0.58\%\\
\hline
Super-2 & 95.24\% $\pm$ 0.96\% & {\bf 95.70}\% $\pm $ 0.33\% & 93.39\% $\pm $ 1.1\% & 94.1\% $\pm $ 0.61\% & 93.11\% $\pm $ 0.5\%\\
\hline
Super-3 & 72.20\% $\pm$ 13.79\% & 77.99\% $\pm$ 12.90\% & {\bf 83.0}\% $\pm$ 6.14\%  & 81.99\% $\pm$ 8.29\% & 64.19\% $\pm$ 8.14\%\\
\hline
Super-4 & 74.57\% $\pm$ 7.23\% & 75.08\% $\pm $ 4.15\% & {\bf 77.16}\% $\pm $ 0.97\%  & 70.95\% $\pm $ 1.86\% & 72.61\% $\pm $ 1.67\% \\
\hline
EBC-1 & 95.17\% $\pm$ 1.21\% & {\bf 95.98}\% $\pm$ 1.28 &93.49$\pm$ 2.66\%  & 91.59\% $\pm$ 4.13\%& 93.96\% $\pm$ 1.1\% \\
\hline
EBC-2 & 84.60\% $\pm$ 7.48\% &{\bf86.73\%$\pm$ 4.03\%} & 83.13\% $\pm$ 3.1\%  & 83.1\% $\pm$ 5.68\% &  86.67\% $\pm$ 1.44\% \\
\hline
EBC-3 & 42.58\% $\pm$ 6.57\% & 45.69\% $\pm$ 10.78\% & 58.62\% $\pm$ 20.37\%  & {\bf 58.7}\% $\pm$ 9.70\%& 52.84\% $\pm$ 5.41\% \\
\hline
\end{tabular}
\end{center}
\label{table:model-performance}
\end{table}

%% The results obtained from the evaluation of each method on the different datasets are presented in Table~\ref{table:model-performance}. For Super-1 and Super-2 datasets, CS-UNet consistently achieves the highest mean IoU scores of 96.14\% and 95.70\%, respectively. UNet and Swin-Unet also perform well on these datasets but exhibit slightly lower accuracies. On Super-3 and Super-4 datasets, Swin-UNet outperforms other methods with mean IoU scores of 83.0\% and 77.16\%, respectively. 

%% EBC-1 dataset showcases comparable performance among the methods, with CS-UNet and UNet achieving the highest mean IoU scores of 95.98\% and 95.17\%, respectively. EBC-2 dataset  CS-UNet and HiFormer leading the pack, followed closely by Swin-Unet and UNet. EBC-3 dataset, characterized by one-shot training images, presents significant challenges. Swin-UNet and TransDeepLabv3+ demonstrate the highest mean IoU scores, although the overall accuracies are considerably lower than in other datasets.

Lastly, we compare the performance of the 3 types of segmentation algorithms (UNet, CS-UNet, Swin-Unet, HiFormer, and TransDeepLabv3+) averaged over different pre-training models and Swin-T architectures in Table~\ref{table:model-performance}. 
The results show that CS-UNet is better than UNet on average across all experiments. While Transformer-based segmentation algorithms may be superior in 1-shot learning or out-of-distribution learning, their performance is not always consistently better than UNet. Therefore, our hybrid algorithm CS-UNet appears to be the more robust solution regardless of the pre-training models.

\section{Conclusion}
\label{sec:conclusion}
In this paper, we demonstrated that training CNN and Transformer encoders on an in-domain dataset can lead to better feature representations for image segmentation.
We introduced CS-UNet -- a hybrid network that combines CNN and Swin Transformer encoders for image segmentation. The encoders of CS-UNet extract low-level and high-level features from input images while the decoder part employs Swin Transformers for image segmentation.
We show that CS-UNet generally outperforms UNet in our experiments and it has competitive performance compared to Transformer-based algorithms. The hybrid nature of CS-UNet enables effective segmentation of images with long-range dependencies and high spatial resolution.

%%We have also introduced a  CS-UNet architecture, a hybrid network combining CNNs and Swin Transformers, for image segmentation. The encoder part of CS-UNet utilizes both CNN and Swin Transformer components to extract low-level and high-level features from input images, respectively. The decoder part employs Swin Transformers for image segmentation. Our comparative analysis of CNN, Transformer, and CS-UNet methods for image segmentation has shown that CS-UNet consistently exhibits better performance across most datasets. UNet also demonstrated good performance and suitability for various contexts. The findings from this study suggest that CS-UNet is a promising approach for image segmentation tasks. It consistently outperformed UNet across all datasets and showed competitive performance compared to Transformer methods in certain scenarios. The hybrid nature of CS-UNet, leveraging both CNNs and Swin Transformers, enables effective segmentation of images with long-range dependencies and high spatial resolution.

Our pre-training dataset is limited in size. A larger dataset will produce better feature representation of microstructures through pre-training, which improves the accuracy of downstream analysis tasks.
While CS-UNet can preserve spatial information and process long-range dependencies, it is computationally intensive compared to CNN and Transformer-based algorithms.
As future work, we will explore the training of other Transformer architectures, such as Focal Transformer and FocalNet, on large microscopy datasets. These architectures may offer further advancements in image segmentation, expanding the range of available options and potential improvements in materials analysis tasks.

%% \section*{Acknowledgement}
%% This research did not receive any specific grant from funding agencies in the public, commercial, or not-for-profit sectors.

\FloatBarrier
%% \newpage

\begin{appendices}
\label{app:sec}

 \section{Transformer Performance}
 \label{sec:Trans}
 Table~\ref{tab:Transformer_acc} shows the best-performing configurations of Transformer-based segmentation algorithms.  
 Note that for pre-training with MicroLite, direct training with MicroLite resulted in the best performance in all cases except Super-4, where ImageNet→MicroLite has the best performance.
 
%For Super-1, both MicroLite and ImageNet models accurately segmented the secondary and tertiary precipitates with an IoU of 95.69\% and 95.56\%, respectively. For Super-2, which contains only four training images, there was a slight reduction in accuracy for both models, to 94.13\% for MicroLite and 94.38\% for ImageNet. For Super-3, which contains only a single training image, the ImageNet model was reduced to 89.78\%, while the MicroLite model was reduced to 92.23\%. The MicroLite model had a high accuracy of 92.23\% IoU during one-shot learning, nearly equal to the accuracy obtained from training on the full dataset. For Super-4, which contains images with different imaging conditions, the MicroLite model performed better than the ImageNet model, with an accuracy of 78.91\% versus 76.42\%.
\begin{table}[h]
\caption{The best performing configurations of Transformer-based segmentation algorithms.}
\label{tab:Transformer_acc}
\begin{center}
\begin{tabular}{|l|ll|ll|}
\hline
\multirow{2}{*}{\textbf{Dataset}} & \multicolumn{2}{l|}{\textbf{ImageNet}} & \multicolumn{2}{l|}{\textbf{MicroLite}} \\ \cline{2-5} 
 & \multicolumn{1}{l|}{\textbf{best performing configuration}} & \textbf{IoU} & \multicolumn{1}{l|}{\textbf{best performing configuration}} & \textbf{IoU} \\ \hline
Super-1 & \multicolumn{1}{l|}{TransDeepLabV3+ / Orginal Swin-T} & 95.25\% & \multicolumn{1}{l|}{Swin-Unet / Intermediate Swin-T} & \bf{95.72\%} \\ \hline
Super-2 & \multicolumn{1}{l|}{Swin-Unet / Orginal Swin-T} & 94.37\% & \multicolumn{1}{l|}{TransDeepLabV3+ / Intermediate Swin-T} & \bf{95.16\%} \\ \hline
Super-3 & \multicolumn{1}{l|}{Swin-Unet / Orginal Swin-T} & 89.78\% & \multicolumn{1}{l|}{TransDeepLabV3+ / Intermediate Swin-T} & \bf{92.23\%} \\ \hline
Super-4 & \multicolumn{1}{l|}{Swin-Unet / Orginal Swin-T} & 76.42\% & \multicolumn{1}{l|}{Swin-Unet / Orginal Swin-T} & \bf{78.91}\% \\ \hline
EBC-1 & \multicolumn{1}{l|}{Swin-Unet / Orginal Swin-T} & 96.11\% & \multicolumn{1}{l|}{TransDeepLabV3+ / Intermediate Swin-T} & \bf{96.59}\% \\ \hline
EBC-2 & \multicolumn{1}{l|}{HiFormer / Orginal Swin-T} & 84.21\% & \multicolumn{1}{l|}{TransDeepLabV3+ / Orginal Swin-T} & \bf{91.11}\% \\ \hline
EBC-3 & \multicolumn{1}{l|}{TransDeepLabV3+ / Orginal Swin-T} & 65.77\% & \multicolumn{1}{l|}{Swin-Unet / Orginal Swin-T} & \bf{82.13}\% \\ \hline
\end{tabular}
\end{center}
\end{table}

\section{Selected CNN Encoders}
\label{app:cnn}

Table~\ref{table:CNNencoders} shows the 19 CNN encoders that were chosen to evaluate the performance of CS-UNet and UNet.

\begin{table}[h]
\caption{Each encoder has at least 1 top-5 average IoU score for Super/EBC datasets (based on Figure 4 of~\cite{stuckner2022microstructure}).} 

\begin{center}
\begin{tabular}{|l|l|l|l|l|l|l|l|}
\hline
CNN Encoder& Super-1 & Super-2 & Super-3 & Super-4 & EBC-1 & EBC-2 & EBC-3 \\ \hline
SE\_ResNet-50~\cite{hu2018squeeze} &  &  &   & &  & & \checkmark
\\ \hline
SE\_ResNeXt-50\_32x4d~\cite{xie2017aggregated} & \checkmark & \checkmark&  & & \checkmark& \checkmark & \checkmark
\\ \hline
SE\_ResNeXt-101\_32x4d~\cite{xie2017aggregated} & \checkmark& \checkmark&  & &  & & \checkmark
\\ \hline
SE\_ResNet-152~\cite{hu2018squeeze} & \checkmark& &  &  & \checkmark&  & 
\\ \hline
SE\_ResNet-101~\cite{hu2018squeeze}& &  & & & \checkmark& \checkmark& \checkmark
\\ \hline
SENet-154~\cite{hu2018squeeze}& & \checkmark & \checkmark& & \checkmark& & 
\\ \hline
ResNeXt-101\_32x8d~\cite{xie2017aggregated}& \checkmark &  & &  &  &  & 
\\ \hline
Inception-V4~\cite{szegedy2017inception}&  & \checkmark & \checkmark&  & \checkmark& &
\\ \hline
Inception-ResNet-V2~\cite{szegedy2017inception}& & & \checkmark& & & &
\\ \hline
DenseNet201~\cite{huang2017densely}& & & & & & \checkmark&
\\ \hline
DenseNet161~\cite{huang2017densely}& & & & & & \checkmark& 
\\ \hline
VGG-16\_bn~\cite{simonyan2014very}& \checkmark & & \checkmark& & & \checkmark&
\\ \hline
VGG-13\_bn~\cite{simonyan2014very}& & & \checkmark&  & &  &
\\ \hline
MobileNet-V2~\cite{sandler2018mobilenetv2}& & & & & & & \checkmark
\\ \hline
EfficientNet-b1~\cite{tan2019efficientnet}& &  & & \checkmark & & &
\\ \hline
EfficientNet-b2~\cite{tan2019efficientnet}& & & & \checkmark& & &
\\ \hline
EfficientNet-b3~\cite{tan2019efficientnet}& & & & \checkmark& & &
\\ \hline
EfficientNet-b4~\cite{tan2019efficientnet}& & & & \checkmark & & &
\\ \hline
EfficientNet-b5~\cite{tan2019efficientnet}& & \checkmark & & \checkmark& & &
\\ \hline
\end{tabular}
\end{center}
\label{table:CNNencoders}
\end{table}

\section{UNet Performance}
 \label{app:Unet}

Table~\ref{table:UNet_acc} shows the CNN encoders that produced the best performance of UNet for each dataset when pre-trained on ImageNet or MicroNet.
EfficientNet and Se\_ResNet families have better performance than the older encoder families such as VGG.

\begin{table}[h]
\caption{The best performing CNN encoders of UNet that are pre-trained on ImageNet and MicroNet.}
\label{table:UNet_acc}
\begin{center} 
\begin{tabular}{|l|ll|ll|}
\hline
\multirow{2}{*}{\textbf{Dataset}} & \multicolumn{2}{l|}{\textbf{ImageNet}} & \multicolumn{2}{l|}{\textbf{MicroNet}} \\ \cline{2-5} 
 & \multicolumn{1}{l|}{\textbf{best performing CNN encoder}} & \textbf{IoU} & \multicolumn{1}{l|}{\textbf{best performing CNN encoder}} & \textbf{IoU} \\ \hline
Super-1 & \multicolumn{1}{l|}{SE\_ResNeXt-50\_32x4} & 96.2\% & \multicolumn{1}{l|}{SE\_ResNeXt-101\_32x4d} & \bf{95.95\%} \\ \hline
Super-2 & \multicolumn{1}{l|}{EfficientNet-b5} & 95.45\% & \multicolumn{1}{l|}{SENet-154} & \bf{95.50\%} \\ \hline
Super-3 & \multicolumn{1}{l|}{EfficientNet-b3} & 70.74\% & \multicolumn{1}{l|}{EfficientNet-b3} & \bf{92.5\%} \\ \hline
Super-4 & \multicolumn{1}{l|}{EfficientNet-b2} & 77.34\% & \multicolumn{1}{l|}{EfficientNet-b1} & \bf{78.95\%} \\ \hline
EBC-1 & \multicolumn{1}{l|}{SENet-154} & 96.23\% & \multicolumn{1}{l|}{SENet-154} & \bf{96.67}\% \\ \hline
EBC-2 & \multicolumn{1}{l|}{DenseNet201} & \bf{91.36}\% & \multicolumn{1}{l|}{InceptionResnetV2} & 90.97\% \\ \hline
EBC-3 & \multicolumn{1}{l|}{EfficientNet-b3} & 44.4\% & \multicolumn{1}{l|}{EfficientNet-b3} & \bf{52.84\%} \\ \hline
\end{tabular}
\end{center}
\end{table}
\FloatBarrier
%In conclusion, the results show that the method pre-trained on MicroLite achieves better segmentation on a one-shot dataset than the method pre-trained on ImageNet under IoU accuracy. Our results also show that the methods pre-trained on MicroLite slightly improve the results on few-shot and out-of-distribution datasets compared to the methods pre-trained on ImageNet.
\section*{DATA AVAILABILITY}
The pre-trained Swin-T models of our experiments are available at our GitHub repository: 

\url{https://github.com/Kalrfou/SwinT-pretrained-microscopy-models}.

This work also used the pre-trained CNN models from 

\url{https://github.com/nasa/pretrained-microscopy-models}.

%The code necessary to apply this technique will be made available on our GitHub repository after the paper is accepted.

\section*{AUTHOR CONTRIBUTIONS}
Alrfou developed the CS-UNet method, conceived and designed the study, developed the software, evaluated results, provided datasets, and contributed to the formal analysis and writing of the original draft. Zhao and Kordijazi evaluated the results, contributed to the formal analysis and writing of the original draft, and proofread and reviewed the final manuscript. 

\end{appendices}
\bibliographystyle{IEEEtran}
\bibliography{main.bib}
%% \begin{appendices}
%% \section{ConvLSTM results}
%% \end{appendices}

%% \newpage
\addtolength{\textheight}{-12cm}   % This command serves to balance the column lengths

%%%%%%%%%%%%%%%%%%%%%%%%%%%%%%%%%%%%%%%%%%%%%%%%%%%%%%%%%%%%%%%%%%%%%%%%%%%%%%%%

%%%%%%%%%%%%%%%%%%%%%%%%%%%%%%%%%%%%%%%%%%%%%%%%%%%%%%%%%%%%%%%%%%%%%%%%%%%%%%%%

%%%%%%%%%%%%%%%%%%%%%%%%%%%%%%%%%%%%%%%%%%%%%%%%%%%%%%%%%%%%%%%%%%%%%%%%%%%%%%%%

%%%%%%%%%%%%%%%%%%%%%%%%%%%%%%%%%%%%%%%%%%%%%%%%%%%%%%%%%%%%%%%%%%%%%%%%%%%%%%%%

\end{document}